\documentclass[journal]{IEEEtran}
\IEEEoverridecommandlockouts
\usepackage{cite}
\usepackage{amsmath,amssymb,amsfonts}
\usepackage{graphicx}
\usepackage{textcomp}
\usepackage{xcolor}
\usepackage[noend]{algpseudocode}
\usepackage{algorithm}
\usepackage{threeparttable}
\usepackage{booktabs}
\usepackage{multirow} 
\usepackage{diagbox}
\usepackage{stfloats}
\usepackage{float} 
\usepackage{subfigure}
\usepackage{epstopdf}
\usepackage{tikz}
\usepackage{subcaption}

\usepackage{url} % to cite web url in the bibtex

\begin{document}

\title{Better YOLO with Attention-Augmented Network\\
and Enhanced Generalization Performance for 
Safety Helmet Detection
}

\author{\IEEEauthorblockN{Shuqi Shen\thanks{Shuqi Shen is with the School Of Automation And Electeical Engineering, University of Science and Technology Beijing, Beijing, China. (email: u202141021@xs.ustb.edu.cn)}}
% \IEEEauthorblockA{\textit{School Of Automation And Electeical Engineering} \\
% \textit{University of Science and Technology Beijing}
% Beijing, China \\
% u202141021@xs.ustb.edu.cn}
% \and
,
\IEEEauthorblockN{Junjie Yang\thanks{Junjie Yang is with the Intelligent Transportation Thrust, Systems Hub, the Hong Kong University of Science and Technology (Guangzhou), Guangzhou, Guangdong, China. (email: junjieyang@hkust-gz.edu.cn)}}
% \IEEEauthorblockA{\textit{Intelligent Transportation Thrust} \\
% \textit{Hong Kong University of Science and Technology (Guangzhou)}\\
% Guangzhou,Guangdong, China \\
% junjieyang@hkust-gz.edu.cn}
}

\maketitle
% \thanks{Shuqi Shen, Junjie Yang}
% \thanks{Shuqi Shen, junjie Yang are with the Intelligent Transportation Thrust, Systems Hub, the Hong Kong University of Science and Technology (Guangzhou), Guangzhou, Guangdong, China. (emails: shuqishen@hkust-gz.edu.cn; junjieyang@hkust-gz.edu.cn)}
% \thanks{Corresponding author: Shuqi Shen (email: shuqishen@hkust-gz.edu.cn)}

\begin{abstract}
Safety helmets play a crucial role in protecting workers from head injuries in construction sites, where potential hazards are prevalent. However, currently, there is no approach that can simultaneously achieve both model accuracy and performance in complex environments. In this study, we utilized a Yolo-based model for safety helmet detection, achieved a 2\% improvement in mAP (mean Average Precision) performance while reducing parameters and Flops count by over 25\%. YOLO(You Only Look Once) is a widely used, high-performance, lightweight model architecture that is well suited for complex environments.  We presents a novel approach by incorporating a lightweight feature extraction network backbone based on GhostNetv2, integrating attention modules such as Spatial Channel-wise Attention Net(SCNet) and Coordination Attention Net(CANet), and adopting the Gradient Norm Aware optimizer (GAM) for improved generalization ability. In safety-critical environments, the accurate detection and speed of safety helmets plays a pivotal role in preventing occupational hazards and ensuring compliance with safety protocols. This work addresses the pressing need for robust and efficient helmet detection methods, offering a comprehensive framework that not only enhances accuracy but also improves the adaptability of detection models to real-world conditions. Our experimental results underscore the synergistic effects of GhostNetv2, attention modules, and the GAM optimizer, presenting a compelling solution for safety helmet detection that achieves superior performance in terms of accuracy, generalization, and efficiency. 
\end{abstract}

\begin{IEEEkeywords}
Compute Vision, Object Detection, Attention mechanism, You Only Look Once (YOLO), Gradient Norm Aware(GAM).
\end{IEEEkeywords}

\section{Introduction}

In various industrial and construction settings, the proper use of safety helmets is fundamental to the well-being of workers. Accurate and efficient detection of safety helmets plays a pivotal role in ensuring occupational safety and compliance with safety protocols. The traditional methods of manual inspection and surveillance are often time-consuming, prone to errors, and insufficient for large-scale operations. In response to these challenges, computer vision and deep learning techniques have emerged as promising tools for automating the detection of safety helmets \cite{patel2023safety,zhao2022improved} .

While existing object detection methods have made significant strides in various domains, safety helmet detection remains a challenging task due to several inherent limitations \cite{yung2022safety,li2022toward}. Traditional methods relying on manual inspection and rule-based systems often fall short in addressing the complexities of dynamic work environments. These methods are prone to false positives and negatives, leading to suboptimal safety enforcement and consequently an increased risk of occupational accidents. Motivated by the overarching goal of bridging the existing gaps in safety helmet detection, we offer a solution that addresses the limitations of current methods. By leveraging the capabilities of GhostNetv2 \cite{22}, attention modules, and the GAM optimizer within the YOLOv5 architecture, we aim to deliver a novel and effective approach to safety helmet detection, contributing to safer and more efficient workplaces.

The foundation of our approach lies in the renowned You Only Look Once (YOLO) algorithm, a real-time object detection system known for its speed and accuracy. YOLO divides an image into a grid, and each grid cell predicts bounding boxes and class probabilities \cite{megalingam2021concurrent,tesema2020denseyolo}. We leverage the YOLOv5 variant for its flexibility and ease of integration to adapt to the conditions of safety helmet detection.

In the quest for heightened object detection precision, we introduce attention mechanisms. Attention mechanisms have garnered significant attention in the field of deep learning for their ability to selectively focus on relevant parts of the input data, thereby enhancing the model's performance in various tasks \cite{niu2021review,shi2022attention}. In the context of object detection, attention mechanisms offer valuable enhancements to convolutional neural networks (CNNs) by allowing the model to dynamically adjust its focus on different spatial regions, features, or channels based on their importance. Incorporating attention mechanisms such as SCNet \cite{zhu2023scnet} and Coordinate Attention \cite{wang2023enhancing} into the YOLOv5 architecture allows us to leverage the benefits of attention-based feature selection while mitigating potential drawbacks. By carefully designing and integrating these attention mechanisms, we aim to enhance the model's ability to capture spatial relationships and improve object detection performance in safety helmet detection tasks.

Achieving robust generalization is challenging due to variations in lighting, backgrounds, and object poses. Our choice to implement the Gradient Norm Aware optimizer (GAM) \cite{20} stems from its ability to smooth the optimization landscape, promoting faster convergence and increased generalization. Improved generalization mitigates overfitting, ensuring the model's effectiveness on new and unseen data.

The primary contributions of our study are outlined as follows:
\begin{enumerate}

\item We tackle the challenge of safety helmet detection by proposing a novel framework, integrating YOLOv5 with attention mechanisms and GhostNetv2-based backbone. This innovation results in a highly efficient model that significantly reduces parameters while maintaining competitive Mean Average Precision (mAP) for accurate safety helmet localization and categorization

\item Unlike prior approaches overlooking global contextual information, our study introduces attention mechanisms, specifically Spatial Channel-wise Attention (SCNet) and Coordinate Attention, within the YOLOv5 architecture. This integration captures both global and local features, enhancing the model's ability to accurately detect safety helmets. 
\item Our study not only introduces innovative model architectures but also prioritizes the crucial aspect of generalization. By carefully designing various model variants with different widths and depths, our approach ensures adaptability to diverse safety helmet detection scenarios. Experimental evaluations conducted on specific safety helmet datasets and commonly benchmarked datasets consistently showcase the improved generalization capabilities of our models.

\end{enumerate}

\section{Related Works}

\subsection{YOLO Architecture}
In the field of object detection, the YOLO (You Only Look Once) \cite{1} series of algorithms and their improved versions have consistently been a focal point of research. Numerous efforts have been dedicated to optimizing and innovating various aspects of object detection based on the YOLO framework. Building upon YOLOv4, YOLOv5 incorporates several improvements, including three data augmentation techniques during data loading, combining CSPNet, Leaky ReLU, and Sigmoid activation functions in the Backbone, and integrating SPP-Net and FPN+PAN structures in the Neck section. Additionally, YOLOv5 introduces adaptive anchor boxes, improving the model's convergence speed and generalization capability. In the field of autonomous driving perception, MCS-YOLO\cite{2} has designed a multi-scale small object detection structure to improve recognition sensitivity and overcome the inherent challenges of detecting small objects. YOLO-Z\cite{3}: Focuses on enhancing small object detection capabilities for autonomous vehicles. By introducing new attention mechanisms and feature fusion strategies, YOLO-Z significantly improves small object detection accuracy while maintaining high speed. As an innovative version of the YOLO series, YOLOX \cite{4}discards the traditional anchor box concept. Through end-to-end training, it simplifies the target detection process and enhances the model's generalization capability. LF-YOLO\cite{5} combines the Reinforced Multi-scale Features (RMF) module to effectively extract multi-scale information using a combination of both parameter-based and parameter-free operations.

\subsection{Attention Mechanisms in Object Detection}
Attention mechanisms\cite{6} have emerged as crucial components in enhancing neural network capabilities, particularly in object detection.  non-local neural networks\cite{7} incorporates non-local operations to capture long-range dependencies in images. This mechanism enables the network to better understand the global context of a scene in object detection tasks. Gated Attention Networks\cite{8} incorporates a gating mechanism to dynamically adjust attention weights in feature maps. This dynamic adjustment enhances the accuracy and robustness of object detection. Cross-Stage Partial Connection\cite{9,10} establishes partial connections between different stages and introducing a novel attention mechanism to enhance object detection performance in few-shot learning scenarios. Multi-Scale Dilated Attention\cite{11,12} employ a multi-head design to execute sliding window\cite{13} with different dilation rates in different heads. CBAM\cite{13} is a well-known attention module like SENet\cite{15}, which combines both spatial attention and channel attention, significantly enhancing the accuracy of the network.

\subsection{Gradient Norm Aware Optimizer}
Optimizer\cite{16,shami2022particle} play a pivotal role in training deep learning models effectively. Stochastic Gradient Descent (SGD)\cite{17,manataki2021comparing} is a fundamental optimization algorithm widely employed in machine learning and deep learning. It minimizes the objective function by iteratively updating model parameters using the gradient of the loss function with respect to those parameters. Adaptive Moment Estimation (Adam)\cite{18} is an adaptive learning rate optimization algorithm that combines ideas from both momentum and RMSprop. It adapts the learning rates for each parameter individually based on the historical gradients, providing efficient and adaptive optimization. Sharpness-Aware Minimization (SAM)\cite{19} is a recent optimization algorithm designed to enhance the flatness of the loss landscape during training. It addresses the issue of overfitting by penalizing sharpness, i.e., the norm of the gradient with respect to the parameters, which sets the stage for the subsequent presentation of the GAM optimizer's\cite{20} integration and its impact on the proposed enhancements.

\section{Methodology}

The structure of our proposed frame work and original YOLOv5 is shown in Fig.\ref{framework}. Compared to the original YOLOv5, this paper has made contributions in the following three aspects: 

\begin{figure*}[htb]
  \centering
  \includegraphics[width=0.7\textwidth]{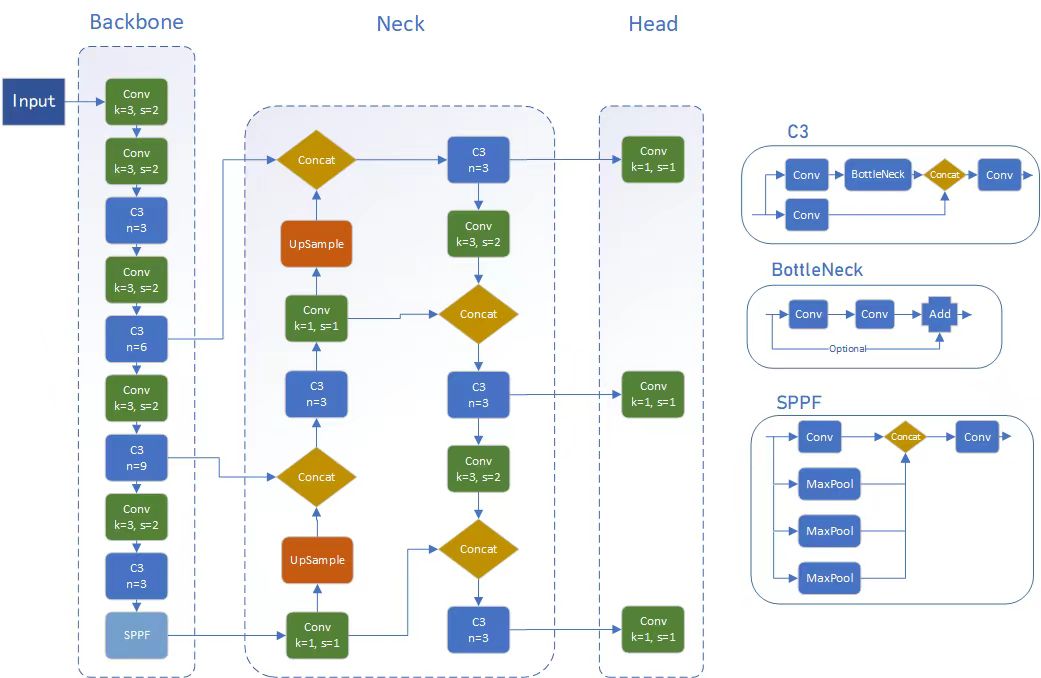}  \includegraphics[width=0.7\textwidth]{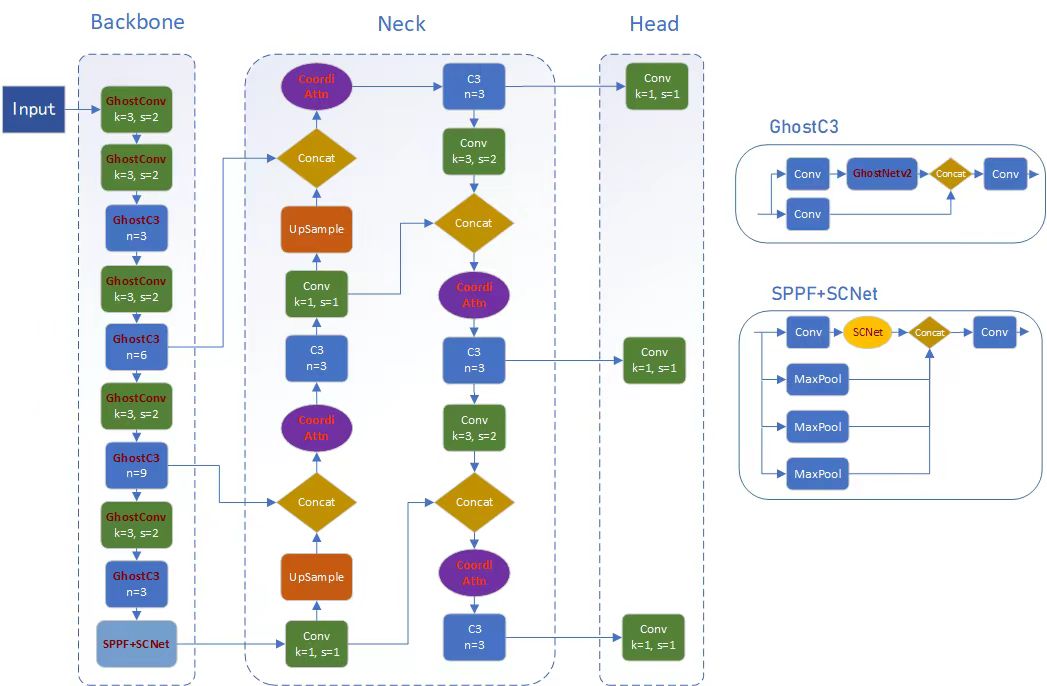}
  \caption{This fig compares our improved framework with the original YOLOv5 framework. In the backbone section, we have replaced the conv and C3 modules with GhostConv and GhostC3, respectively. Additionally, we have substituted the original SPPF with an SPPF integrated with SCNet. In the neck section, we have added a Coordination and Attention (CA) module following each concatenation step.}
  \label{framework}
\end{figure*}

\begin{enumerate}
\item Substituting the original backbone with GhostNetv2 to efficiently extract feature maps with reduced parameter complexity. 
\item Introducing attention mechanisms, specifically Self-Calibrated Convolutions and Coordinate Attention, within both the backbone and neck components, enabling the model to better focus on relevant information and improve accuracy when dealing in various environment.
\item Adding the Gradient Norm Aware optimizer to the original optimization method, enhancing model’s generalization capability.
\end{enumerate}

\subsection{Lightweight Feature Extraction Network Backbone Based on GhostNetV2}
\begin{figure}[h]
  \centering  \includegraphics[width=0.5\textwidth]{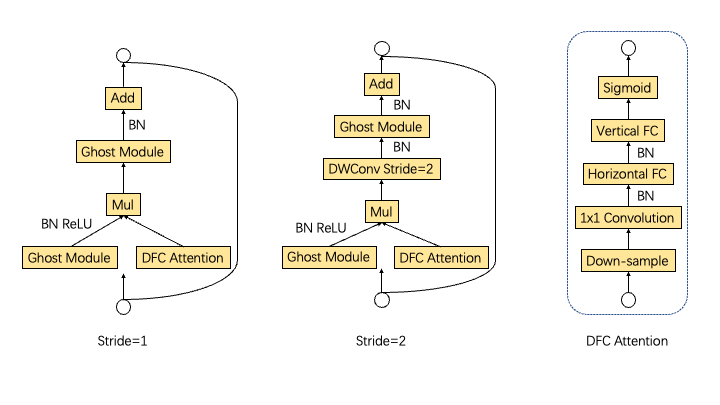}
  \caption{The framework of the GhostNetV2 Bottleneck}
  \label{threeframes}
\end{figure}

The C3 module in the backbone introduces complex structures and connectivity methods, which may lead to redundant parameters. Therefore, GhostNet\cite{21,22} introduces Depth-wise convolution and point-wise convolution to reduce parameters. Additionally, it incorporates DFC attention\cite{23} based on fully connected layers to address the issue of small convolutional local receptive fields. GhostNetV2 emphasizes parameter efficiency through Ghost blocks and excels at feature extraction in deep neural networks. Its design prioritizes performance without compromising computational resources. The network backbone of our frame work is shown in Fig.\ref{threeframes}

\subsubsection{GhostNet Module}
 A standard GhostNet module can replace a conventional convolution block for an input feature \(X \in \mathbb{R}^{C\times H\times W}\) in following steps: First, utilizing deep convolution to reduce the channel dimension of input feature maps, with the outcome denoted as \(Y'\)\eqref{eq1}. Then, employing Point-wise convolution to capture channel information, followed by concatenation with \(Y'\)\eqref{eq2}. Next, Multiplying with DFC Attention to capture spatial neighborhood information, where DFC Attention map can be computed as \eqref{eq3}., \(Z\) is the \(HW\) tokens \(\in \mathbb{R}^{C}\).

\begin{equation}
Y'=X*F_{1\times1}\label{eq1}
\end{equation}

\begin{equation}
Y=Concat([Y', Y'*F_{dp}])\label{eq2}
\end{equation}

\begin{equation}
a'_{hw}=\sum^{H} F_{h,h'w}^H \odot z_{h'w} \label{eq3}
\end{equation}
\[a_{hw}=\sum_{w'=1}^W F_{w,hw'}^W \odot a'_{hw'}\]

By independently applying convolution operations to the Height and Width dimensions, DFC Attention significantly reduces parameters and computational load while extracting long-range features. Under the same conditions of input feature maps \(\mathbb{R}^{C\times H\times W}\), and the same corresponding output \(\mathbb{R'}^{C'\times H\times W}\), the total parameters for a standard convolution module and GhostNet module can be calculated as follows:

\begin{equation}
P_{conv}=C\cdot C'\cdot k^2+2\cdot C'\label{eq4}
\end{equation}
\[P_{Ghost}=(C\cdot C_{mid}\cdot k^2 + 2 C_{mid})\cdot 2+C_{mid}\cdot C'\cdot k^2+2\cdot C'\]

\(C_{mid}\) is usually half the size of the input channel. It's obvious that the ghost module needs fewer parameters compared to a standard convolution module. 

\subsubsection{GhostNet Bottleneck}
The GhostNet bottleneck consists of two GhostNet modules. The first Ghost module extracts deeper information and enriches features, while the subsequent Ghost module, without DFC attention, changes the number of channels to match the residual path. The structure of Ghost Bottleneck is shown in below:    
% Fig
Due to the aforementioned advantages of GhostNet, after replacing the C3 and Conv modules in the backbone with GhostNet, the computational and parameter sizes were significantly reduced, improving the operating speed of the network without a noticeable loss in accuracy. 

\subsection{Attention Modules Integration to Compensate for Accuracy}
While GhostNet markedly reduces model parameters and computational load, it struggles to effectively capture spatial features, inevitably leading to some accuracy loss. To compensate, we propose attention mechanisms, especially Self-Calibrated Convolutions and Coordinate Attention in YOLO's backbone and neck sections, preserving long-range positional information. The attention mechanism enables the model to selectively focus on certain parts of the input sequence or image by learning weights to allocate attention to different positions. This allows the model to selectively attend to specific regions of the input while ignoring other noise or irrelevant parts.

\subsubsection{Self-Calibrated Convolutions (SCNet)}
\begin{figure}[h]
\centering
  \includegraphics[width=0.4\textwidth]{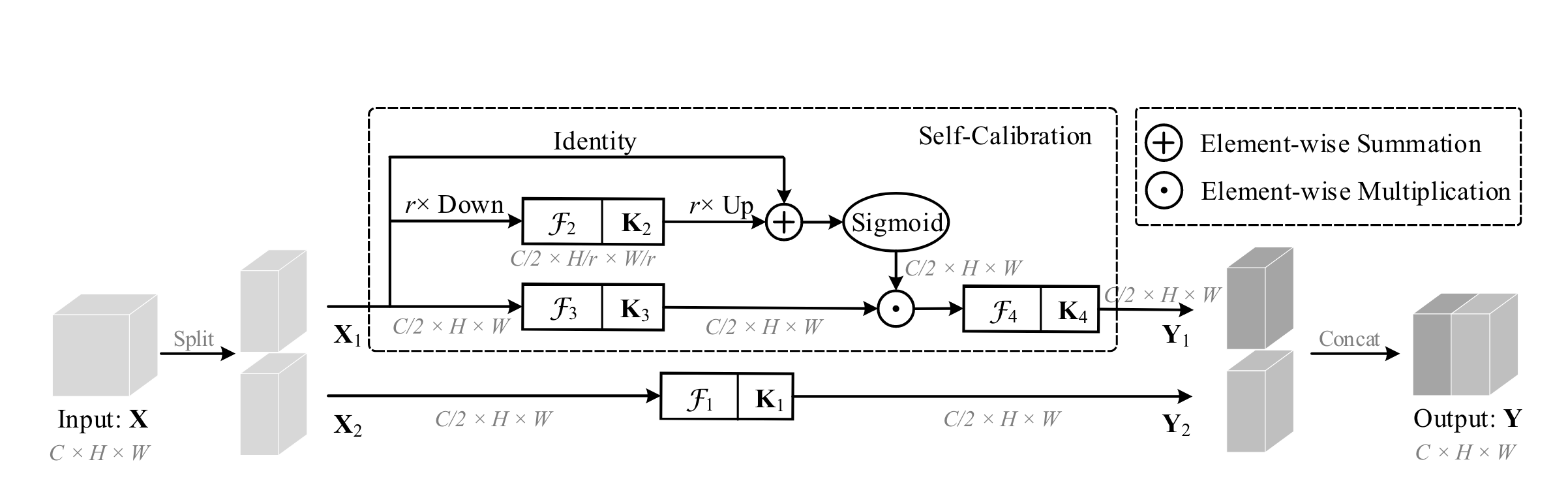}
  \caption{The workflow of SCNet}
  \label{scnet}
\end{figure}

SCNet is a lightweight attention module, the workflow is shown in Fig.\ref{scnet} In SCNet, the input is split into two branches. They are computed individually and the result of each branch is concatenated to the final result.

As the figure above shows, SCNet has several different sizes of residual block that guide the output to highlight the region of interest. This embedding of parallel branches also provides the module with a large fields-of-view, which helps it learn its spatial surroundings. With the multi-scale features, the model can effectively detect small helmet targets in the various environment. 
In our proposed module, SCNet is added in the main branch of SPPF block, which is shown in Fig7. As also a residual block, but different from SCNet, SPPF block concatenates unequal channel size of feature, while SCNet connects different scales of resolution. By adding SCNet in SPPF block, SPPF will learn feature relationships between different resolutions, strength the ability to detect objects of different sizes. Moreover, the shortcoming of SCNet, the weak ability to capture large receptive fields due to a stack of such convolution layers, is unexposed to the maximum extent. 

\subsubsection{Coordinate Attention (CA)}
\begin{figure}[h]
  \includegraphics[width=0.4\textwidth]{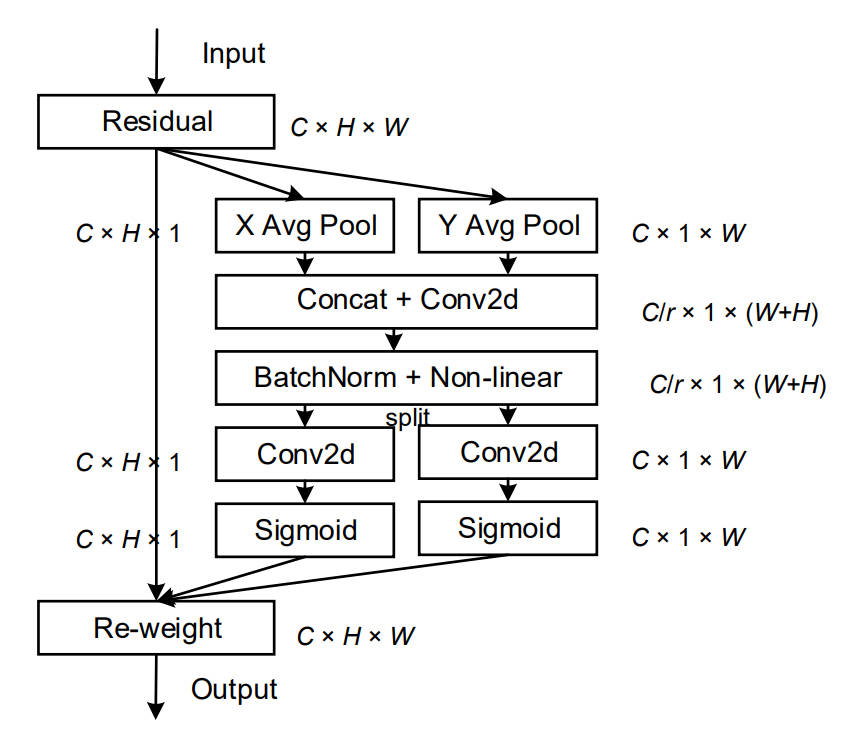}
  \caption{The schematic diagram of the CA module}
  \label{ca}
\end{figure}

The Neck in YOLO plays a crucial role in decoupling and extracting features from the feature maps. In order to address the issue of losing precise spatial relationships of inputs during the upsampling process, Coordinate Attention is proposed. CA enhances feature learning by incorporating spatial relationships based on pixel coordinates. The schematic diagram of the CA module is shown in Fig.\ref{ca}.

As illustration shows, CA module consists of 2 step: coordinate information embedding and coordinate attention generation. 
coordinate information embedding used global pooling to embed spatial features. For input feature \(\mathbb{R}^{C\times H\times W}\), two pool kernel \((1,W),(H,1)\)are separately implemented to \(W\) \eqref{eq5} and \(H\) \eqref{eq6} dimension

\begin{equation}
z_c^h(h)=\frac{1}{W}\sum_{0\leq i\geq W}x_c(h,i)\label{eq5}
\end{equation}

\begin{equation}
z_c^w(w)=\frac{1}{H}\sum_{0\leq j\geq H}x_c(j,w)\label{eq6}
\end{equation}

The results \(z_h\) and \(z_w\) represent relationship between pixels and their cross neighborhood, but not have precise positional information, which terms coordinate attention generation. Thus another two \(1\times 1\) convolution transformations was added to \(f^h\) and \(f^w\)\eqref{eq8}, where \(f\) was the concatenation output of \(z^h\) and \(z^w\)\eqref{eq7}.

\begin{equation}
f = \delta(F_1([z^h,z^w]))\label{eq7}
\end{equation}

\begin{equation}
g^h = \sigma(F_h(f^h))\label{eq8}
\end{equation}
\[g^w = \sigma(F_w(f^w))\]

The framework of the proposed Neck with Coordinate attention is shown in Fig.\ref{framework}. In our module, the CA module is added to front of the C3 block in Yolo’s Neck out of two concerning: 1. We believe GhostNet has effectively encoded different size of features from down to top, and C3 blocks in Neck do further integrate and enhance features. With CA block reweighting pixels’ position relationship, Neck will focus more on the most useful information. 2. With 4 different C3 blocks handing features of different sizes and depths in Neck, adding CA in front of C3 blocks will maximizing the utilization of position information and improving precision. 

\subsection{GAM Optimizer Implementation}
Detecting safety helmets is performed in complex and diverse scenarios. Thus enhancing the model's generalization capability is essential. Based on Stochastic Gradient Descent (SGD), we added first-order flatness on optimizer to improve generalization performance.
SGD employs stochastic approximation techniques to minimize the loss function by adjusting model parameters based on randomly sampled subsets of training data, facilitating the convergence of the model towards the optimal solution. The SGD algorithm can be represented by the following formula\eqref{eq9}, where \(\eta\) is learning rate, \(t\) is time step, \(g_t\) is the loss gradient.

\begin{equation}
\theta_{t+1}=\theta_t-\eta g_t\label{eq9}
\end{equation}

To enhance the model's generalization capability, we added GAM (Gradient norm aware minimization) to SGD optimizer. The maximum eigenvalue of the Hessian is widely recognized as an indicator for measuring the smoothness and curvature of the convergence position. GAM approximates the maximum eigenvalue of the Hessian using the following formula\eqref{eq10}:

\begin{equation}
\lambda_{\max} \left( \nabla^2 \hat{L}(\theta^*) \right) = \frac{R_{\rho}^{(1)}(\theta^*)}{\rho^2}\label{eq10}
\end{equation}

Where \(R_{\rho}\) is the first-order flatness, can be computed in \eqref{eq11}:

\begin{equation}
R^{(0)}_{\rho} (\theta) \triangleq \max_{\theta' \in \mathcal{B}(\theta, \rho)} \left( \hat{L}(\theta') - \hat{L}(\theta) \right), \quad \forall \theta \in \Theta\label{eq11}
\end{equation}

GAM constrains the range of generalization error using Formula\eqref{eq12} . As the Format shown, the key to optimize generalization error is to control loss function \(\hat{L}(\boldsymbol{\theta})\) and first-order flatness \(R_{\rho}\). By updating weight controlling gradient of loss value and first-order flatness, GAM Gradually reduce generalization errors.

\begin{equation}
\begin{array}{l} 
\mathbb{E}_{\epsilon_{i} \sim N\left(0, \rho^{2} /(\sqrt{d}+\sqrt{\log n})^{2}\right)}[L(\boldsymbol{\theta}+\boldsymbol{\epsilon})] \\
\leq \hat{L}(\boldsymbol{\theta})+R_{\rho}^{(1)}(\boldsymbol{\theta})+\frac{M}{\sqrt{n}} \\
+\sqrt{\frac{\frac{1}{4} d \log \left(1+\frac{\|\boldsymbol{\theta}\|^{2}(\sqrt{d}+\sqrt{\log n})^{2}}{d \rho^{2}}\right)+\frac{1}{4}+\log \frac{n}{\delta}+2 \log (6 n+3 d)}{n-1}} \label{eq12}
\end{array}
\end{equation}

The code flow is as follows:

\begin{algorithm}
\caption{Gradient norm Aware Minimization (GAM)}
\begin{algorithmic}[1]
\State \textbf{Input}: Batch size $b$, Learning rate $\eta_t$, Perturbation radius $\rho_t$, Trade-off coefficient $\alpha$, Small constant $\xi$
\State $t \leftarrow 0$, $\theta_0 \leftarrow \text{initial parameters}$
\While{$\theta_t$ not converged}
    \State Sample $W_t$ from the training data with $b$ instances
    \State $h_t^{\text{loss}} \leftarrow \nabla L^{\text{oracle}}(\theta_t)$ \Comment{Calculate the oracle loss gradient $\nabla L^{\text{oracle}}(\theta_t)$}
    \State $f_t \leftarrow \nabla^2 \hat{L}_{W_t}(\theta_t) \cdot \frac{\nabla \hat{L}_{W_t}(\theta_t)}{\| \nabla \hat{L}_{W_t}(\theta_t) \| + \xi}$
    \State $\theta_{t}^{\text{adv}} \leftarrow \theta_t + \rho_t \cdot \frac{f_t}{\|f_t\| + \xi}$
    \State $h_{t}^{\text{norm}} \leftarrow \rho_t \cdot \nabla^2 \hat{L}_{W_t}(\theta_{t}^{\text{adv}}) \cdot \frac{\nabla \hat{L}_{W_t}(\theta_{t}^{\text{adv}})}{\| \nabla \hat{L}_{W_t}(\theta_{t}^{\text{adv}}) \| + \xi}$ \Comment{Calculate the norm gradient $\nabla R_{\rho_t}^{(1)}(\theta_t)$}
    \State $\theta_{t+1} \leftarrow \theta_t - \eta_t(h_t^{\text{loss}} + \alpha h_{t}^{\text{norm}})$
    \State $t \leftarrow t + 1$
\EndWhile
\State \textbf{return} $\theta_t$
\end{algorithmic}
\end{algorithm}

Therefore, the model converges in the direction of reducing generalization errors. By adding GAM to optimizer, the generalization capability of the module can be effectively enhanced.
\begin{figure*}[htp]
  \centering
  \subfigure[]{
  \includegraphics[width=0.15\textwidth]{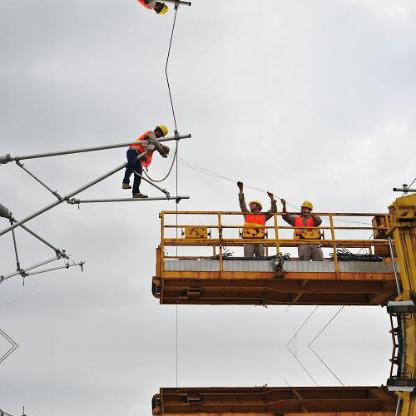}
  \includegraphics[width=0.15\textwidth]{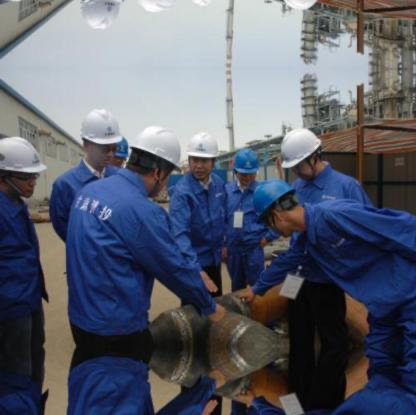}
  \includegraphics[width=0.15\textwidth]{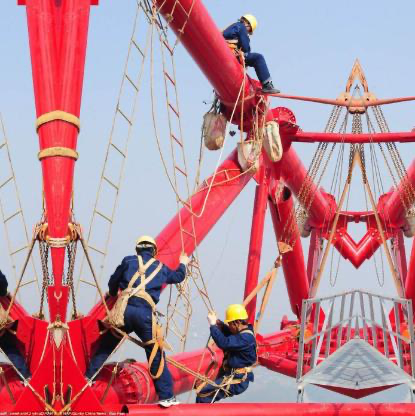}
  \includegraphics[width=0.15\textwidth]{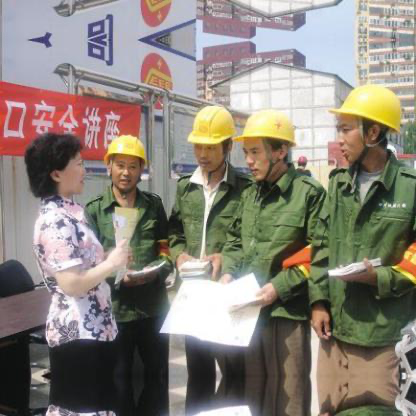}
  }
  \subfigure[]{
  \includegraphics[width=0.15\textwidth]{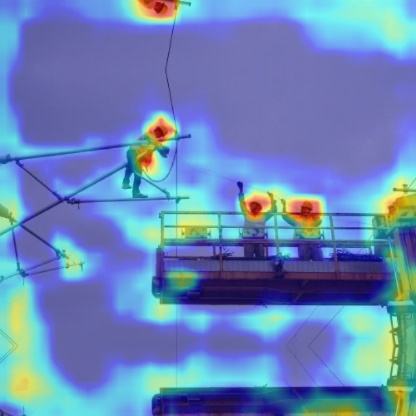}
  \includegraphics[width=0.15\textwidth]{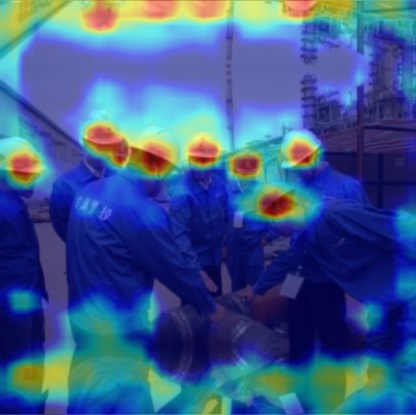}
  \includegraphics[width=0.15\textwidth]{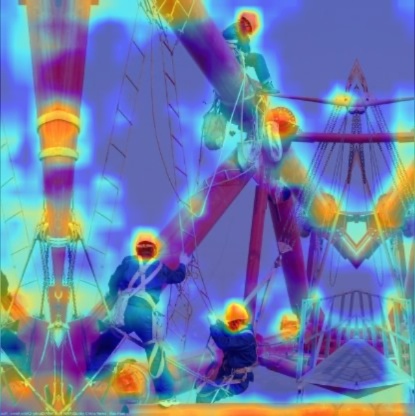} 
  \includegraphics[width=0.15\textwidth]{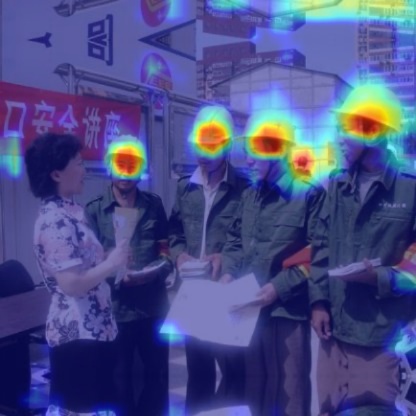}
  }
  \subfigure[]{
  \includegraphics[width=0.15\textwidth]{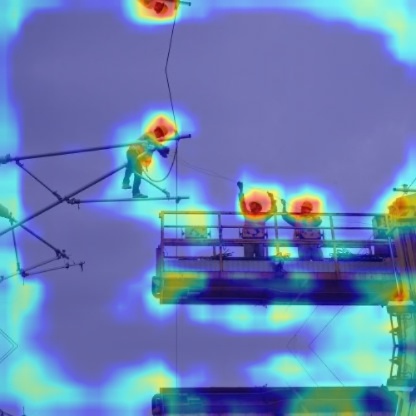}
  \includegraphics[width=0.15\textwidth]{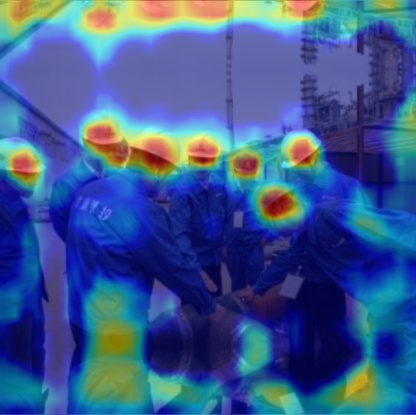}
  \includegraphics[width=0.15\textwidth]{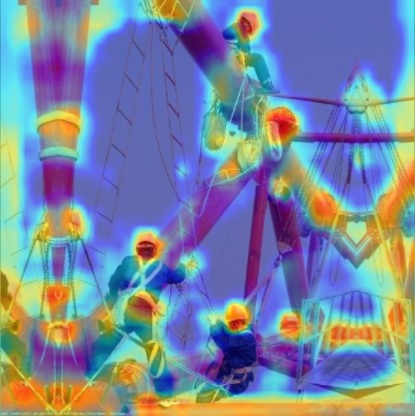} 
  \includegraphics[width=0.15\textwidth]{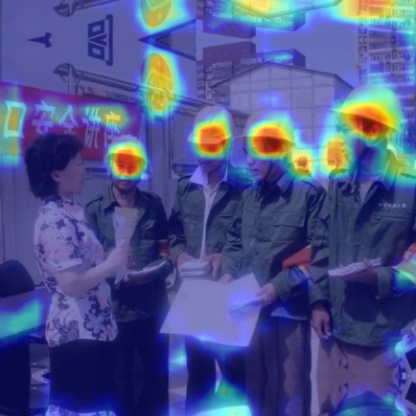}
  }
  \subfigure[]{
  \includegraphics[width=0.15\textwidth]{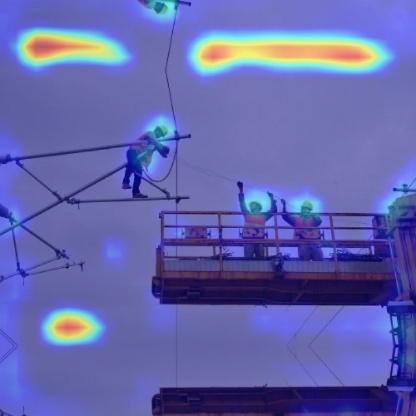}
  \includegraphics[width=0.15\textwidth]{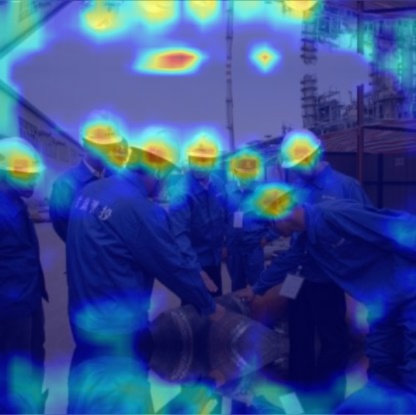}
  \includegraphics[width=0.15\textwidth]{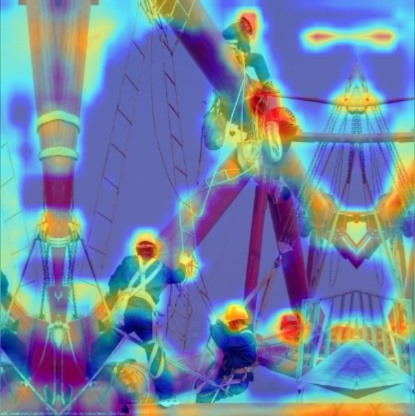} 
  \includegraphics[width=0.15\textwidth]{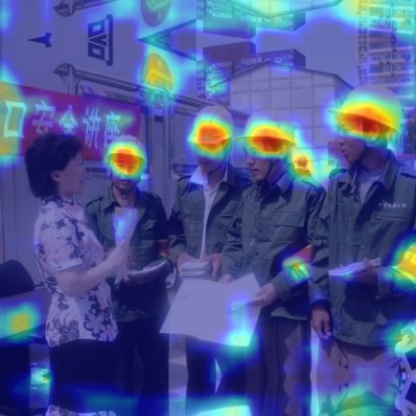}
  }
  \subfigure[]{
  \includegraphics[width=0.15\textwidth]{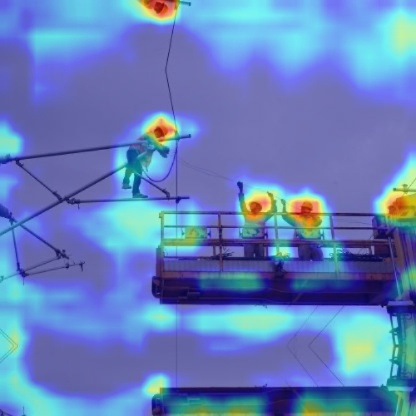}
  \includegraphics[width=0.15\textwidth]{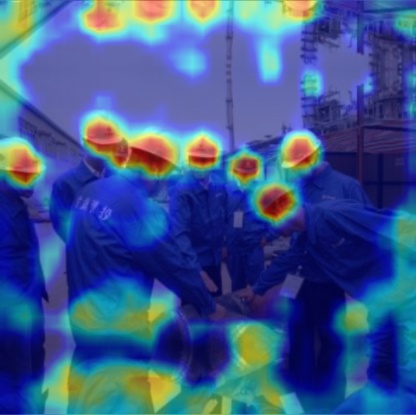}
  \includegraphics[width=0.15\textwidth]{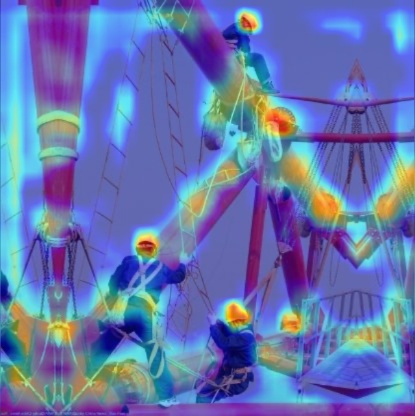} 
  \includegraphics[width=0.15\textwidth]{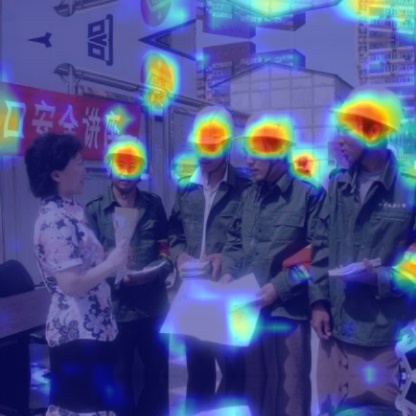}
  }
  \caption{This figure displays the feature maps generated by different experimental networks for the same test image. These four rows experimental images are respectively from experiments 2, 6, 7, and 8 in Table 1.}
  \label{heatmap}
\end{figure*}
\section{Experiments}

In this section, we present a series of experiments designed to validate the proposed enhancements to YOLOv5 for helmet detection. The experiments are conducted on the publicly available helmet detection dataset provided by Kaggle. We comprehensively evaluate both the original YOLOv5 and its variants incorporating various enhancements, conducting thorough comparisons based on key metrics such as mean Average Precision (mAP), parameters, GFLOPs, model size, and Intersection over Union (IoU).

\subsection{Experimental Design}
Our experiments were conducted on a hardware platform equipped with an AMD CPU 5800x and an NVIDIA GeForce RTX4090, utilizing CUDA 11.7 and PyTorch 1.18.1 as the underlying deep learning framework. All datasets, neural network models, and associated resources used in this study are readily accessible and validated within the aforementioned environment.

The dataset employed in our research is sourced from eight Kaggle's open-access helmet detection projects, comprising a total of over 20000 image samples. We partitioned 90\% images for training and 10\% images for validation to assess the effectiveness of our proposed methodologies. Apart from the modifications proposed in this paper, default parameters were maintained for all YOLOv5 training configurations configurations (further details can be elaborated upon).
% add dataset hyperparm table
In addition to the YOLOv5 model, we selected variants such as Faster R-CNN, YOLOv5-L, Yolov3 for comparative analysis against our proposed methods. The comparative experiments show the outstanding performance of our proposed model in terms of detection accuracy, detection speed, and other aspects.

\subsection{Ablation Study and Visualization}
We conducted ablation experiments using the following methods employed in this study: GhostNetV2, Coordinate Attention (CA), Self-Calibrated Convolutions (SCNet), and GAM optimizer. We analyzed the results by individually integrating each of these methods into Yolov5 on the dataset shared on Alibaba Cloud \cite{Data}. Based on the findings from the following results, we arrived at the following conclusions:
% do cite here!!!!!!!
\begin{table}[htp] \label{table1}
	\centering
	\fontsize{6}{10}\selectfont    %{字体尺寸}{行距}
	\caption{Comparison of results between our complete framework and the framework with some modules removed}
	\begin{tabular}{|c|c|c|c|c|c|c|c|}
            \hline
  		\multirow{2}{*}{\#}&\multicolumn{4}{c|}{Method}&\multicolumn{3}{c|}{Evaluate}\\
            % \hline
            \cline{2-8}
		   &\tiny{GhostNet} &\quad CA \quad & SCNet & GAM & mAP(50) & Params(M) & {Size(Mb)}  \\
		\hline    
		      1\label{exp1}& &  &  &  & 94.817 & 7.025 & 14.5 \\
            % \hline 
		    2\label{exp2}&\checkmark&  &  &  & 94.775 & {\bf 5.142} & {\bf 10.5} \\
                3\label{exp3}& &\checkmark&  &  & {\bf 95.626} & 7.520 & 15.5 \\
                4\label{exp4}& &  &\checkmark&  & 94.9 & 7.140 & 14.0 \\
                5\label{exp5}& &  &  &\checkmark& 94.99 & 7.025 & 14.5 \\
                6\label{exp6}& \checkmark&\checkmark&  &  & 95.11 & 5.638 & 11.8 \\
                7\label{exp7}& \checkmark&  &\checkmark&  & 95.147 & 5.256 & 10.8 \\
                8\label{exp8}& \checkmark&\checkmark&\checkmark&  & 95.479 & 5.752 & 12.0 \\
                9\label{exp9}& \checkmark&\checkmark&\checkmark&\checkmark& 95.558 & 5.752 & 12.0 \\
  		\toprule
	\end{tabular}
	\label{tab:Training_sizes}
\end{table}

\renewcommand{\arraystretch}{1.5} %控制行高
\begin{table*}[ht]
	\centering
	\fontsize{7.5}{10}\selectfont
	\begin{threeparttable}
		\caption{Comparison between our proposed framework and other common frameworks}
		\label{tab:perfor}
		\begin{tabular}{ccccccc}
			\toprule
			\toprule
			% \multirow{1}{*}{D}&C
			& Method & Backbone & mAP@0.5(\%) & Parameters(M) & Model Size(Mb)& FPS \cr
			\midrule
			% \multirow{3}*{ES}
			   & Cascade R-CNN (CVPR 2018)\cite{cai2018cascade} & ResNeXt-101 & 92.820 &74.8 & 273 & 4.3  \cr
			~ & Fast R-CNN\cite{girshick2015fast} & FPN & 92.01 & 51.1 & 206 & 9.2  \cr
			% ~ & CornerNet511 & Hourglass-52 &  &  &  &   \cr
			~ & Mask R-CNN\cite{he2017mask} & FPN & 92.87 & 44.8 & 121 & 12  \cr
			~ & SSD300\cite{liu2016ssd} & VGGNet & 91.79 & 26.5 & 79.2 & 44  \cr
			~ & RetinaNet\cite{lin2017focal} & Swin Transformer-T & 95.92 & 37.1 & 103.6 & 18 \cr
			~ & ConvNext\cite{liu2022convnet} & ConvNext & 93.38 & 68.5 & 137.2 & 33  \cr
			~ & YoloV2\cite{redmon2016yolo9000} & Darknet19 & 91.256 & 48.11 & 193.2 & 51  \cr
			~ & YoloV3\cite{redmon2018yolov3} & Darknet53 & 92.214& 62.37 & 237.1 & 46 \cr
			~ & YoloV3-tiny\cite{adarsh2020yolo} & FCCL & 91.301 & 8.255 & 33.8 & 72  \cr
			~ & YoloV5-s & C3 & 94.316 & 7.235 & 14.1 & 72  \cr
			~ & YoloV5-s & CSP & 94.201 & 7.523 & 14.5 & 72  \cr
			~ & YoloV5-x & C3 & 96.544 & 86.479 & 166 & 63  \cr
			~ & YoloV8-s & C2f & 95.218 & 11.14 & 22.5 & 65  \cr
			~ & YoloX-x\cite{ge2021yolox} & Modified CSP & 96.1 & 99.0 & 345 & 22  \cr
			~ & Yolo-Former\cite{dai2022yolo} & C3-Former & 96.2 & 26.73 & 32.5 & 44 \cr
			~ & HIC-Yolo\cite{tang2023hic} & C3 with CBAM & 94.40 & 8.39 & 18.2 & 69 \cr
                
                % need find more
                %yolo with new backbone and module
			\hline
			% \multirow{1}{*}{Run Time (Seconds)}&
			& Better Yolo (ours) & GhostNetv2 &95.558&5.752&12.0&78 \cr
			\bottomrule
			\bottomrule
		\end{tabular}
	\end{threeparttable}
\end{table*}

\begin{figure}[ht]
  \centering
  \includegraphics[width=0.5\textwidth]{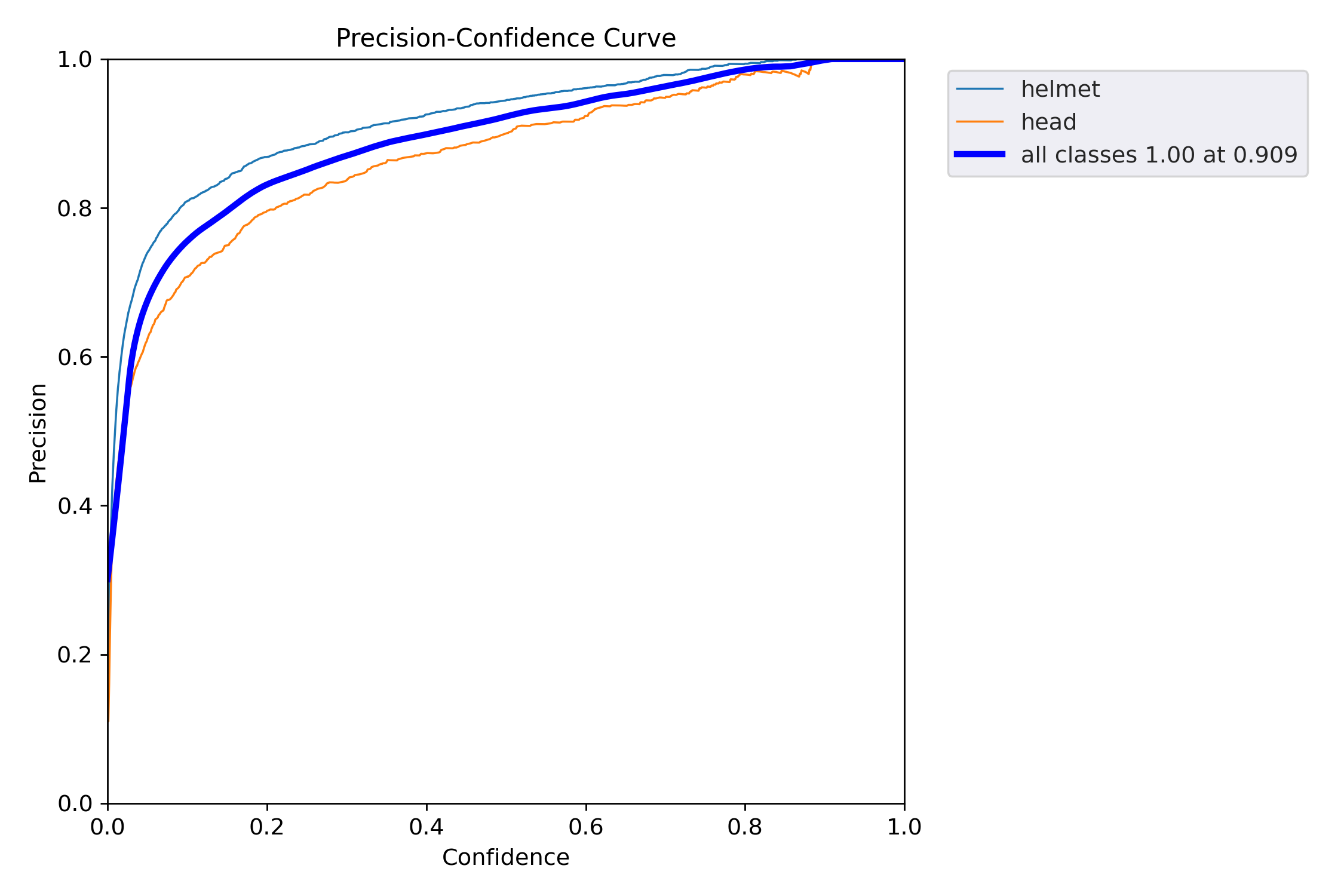}
  \includegraphics[width=0.5\textwidth]{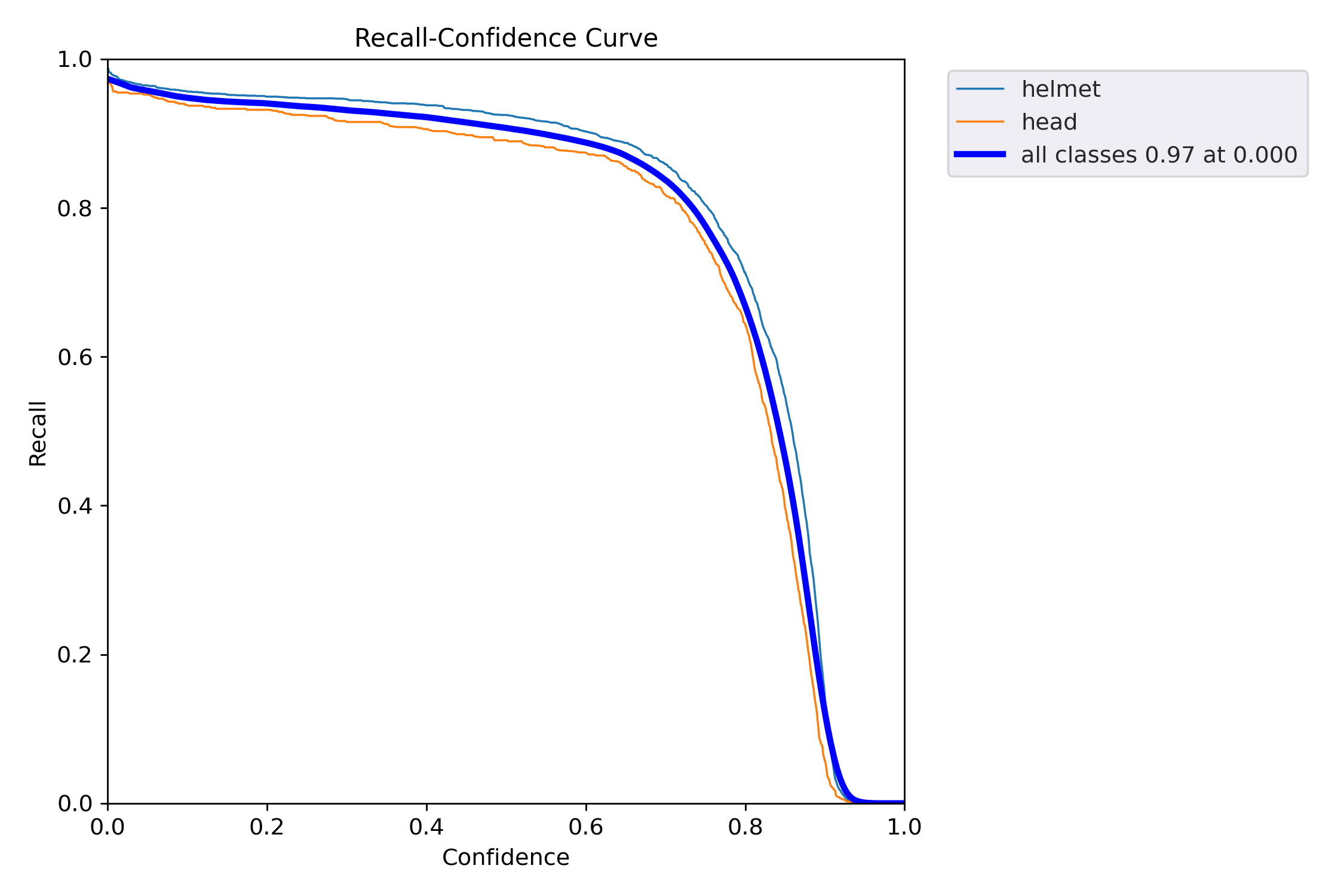}
  \includegraphics[width=0.5\textwidth]{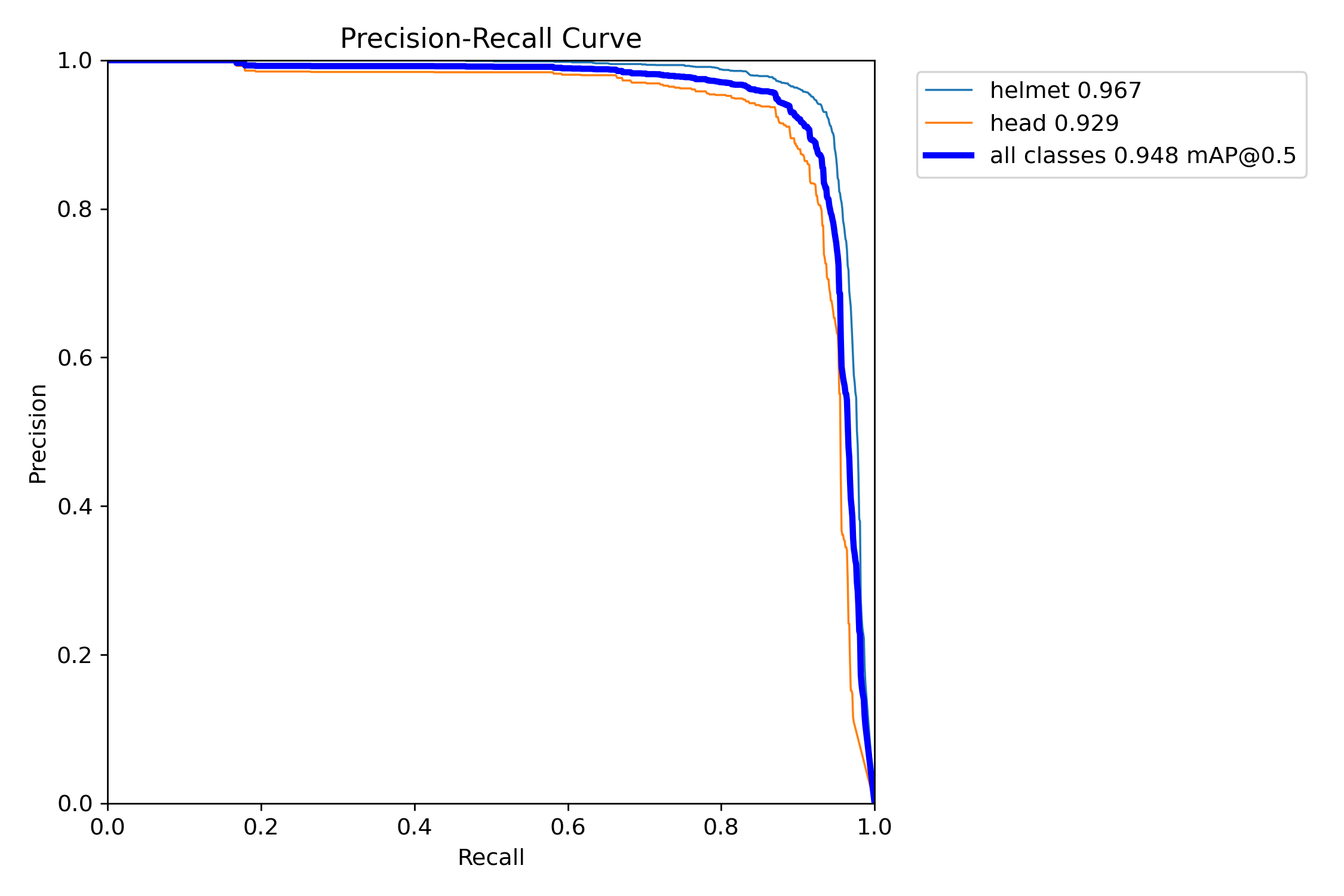}
  \caption{The curve result of precision confidence, precision-recall and recall confidence of our proposed module.}
  \label{curve}
\end{figure}

\begin{enumerate}
\item Yolo combined with CA achieved the highest MAP, but blindly adding attention modules also increased the parameter count, resulting in model redundancy, while employing GhostNet significantly reduced the model parameters by over 25\%. Meanwhile, directly incorporating SCNet does not effectively improve model accuracy, but SCNet's spatial capturing ability can effectively compensate for the accuracy loss caused by reducing model size. In summary of the results, The model we proposed can maintain high accuracy while significantly reducing the parameter count.

\item We selected some test images along with the heat maps generated from each experiment for comparison \ref{heatmap}. In the displayed feature maps, our proposed model demonstrates excellent and consistent feature capturing capabilities for safety helmets overall, maintaining good performance even in complex and interfering environments, specifically demonstrated by paying less attention to interference, doing superior segmentation of targets and surroundings, and so on. 
\item The final result curves is shown in Fig.\ref{curve}. The curve shows that the model has higher precision for detecting "helmet" compared to "head". This could be because most of the head data appearing in the dataset has been subjected by helmets to varying degrees of occlusion. Both classes reach their peak precision at high confidence levels in Precision-Confidence Curve. The area under the P-R curve (AUC) for both classes is quite high, making it reliable for applications where accurate and robust object detection is needed.
\end{enumerate}
% \begin{figure}[ht]
%   \includegraphics[width=0.55\textwidth]{fig/P_curve.png}  
%   \includegraphics[width=0.55\textwidth]{fig/PR_curve.png}  
%   \includegraphics[width=0.55\textwidth]{fig/R_curve.png}  
%   \caption{the explanation}
% \end{figure}

% \begin{table}[h]
% \scriptsize
%    \centering%\multicolumn{3}{ r}
%    \caption{backbone}
%    \label{tab4}
%     \begin{tabular}{c c  c c c}\hline
% 	  Layer & $c_i$ & $k$ & Parameter(k)\\ \hline
%         Conv & 3 & 6 & 3.52\\
%         GhostConv & 32 & 3 & 9.63\\
%         C3Ghost & 64 & 5 & 10.1\\
%         GhostConv & 64 & 3 & 37.7\\
%         C3Ghost & 128 & 5 & 43.7\\
%         GhostConv & 128 & 3 & 149.1\\
%         C3Ghost & 256 & 5 & 188.1\\
%         GhostConv & 256 & 3 & 593.2\\
%         C3Ghost & 512 & 5 & 596.4\\
%         SPPF & 512 & 5 & 771.3\\
%       \hline
%     \end{tabular}
% \end{table}
\subsection{Comparative Experiment}
% lr设置啥的
In this segment, we embarked on a comprehensive exploration through a series of comparative experiments, spanning across various networks. Our primary objective was to meticulously evaluate and appraise the efficacy of our proposed model, meticulously scrutinizing a spectrum of metrics for thorough analysis.
Moreover, we delved deeper into the assessment realm by orchestrating cross-dataset comparisons across multiple datasets in tandem. This strategic maneuver was instrumental in gauging the overarching generalization prowess inherent within our proposed model. By subjecting our model to diverse datasets simultaneously, we aimed to ascertain its adaptability and robustness in disparate scenarios.
To enrich our comparative analysis, we meticulously curated three distinct helmet datasets sourced from Kaggle, meticulously handpicked to encapsulate a diverse range of scenarios and challenges. Through rigorous experimentation, we endeavored to juxtapose the performance of our proposed model against that of YOLOv5s across these varied datasets.
\begin{table}[ht]
\scriptsize
   \centering%\multicolumn{3}{ r}
   \caption{Results of our proposed framework and Yolov5-s on four different datasets}
   \label{tab4}
    \begin{tabular}{c c  c c c}\hline
	Dataset& 1\cite{kaggleData1} & 2\cite{kaggleData2}& 3\cite{kaggleData3}& 4\cite{kaggleData4}\\ \hline
        Yolov5-s &96.201 & 87.010 & 88.236 & 68.582 \\
        Our Model & 96.558 & 87.291 & 91.280 & 72.173 \\
      \hline
		\end{tabular}
\end{table}
% \begin{figure*}[hb]
%   \includegraphics[width=1.0\textwidth]{fig/results22.png}  
%   \caption{the explanation}
% \end{figure*}
\section{Conclusion and Discussion}
We proposed a YOLO-based model characterized by low parameters, high detection accuracy, and robust generalization capabilities, enabling safe operation in complex environments. However, regardless of whether attention mechanisms can significantly improve detection accuracy, the magnitude of improvement is always limited. How to further enhance the model's performance while maintaining its lightweight characteristics is a question that everyone needs to consider.

\bibliographystyle{ieeetr}
\bibliography{Reference}

\begin{IEEEbiography}[{\includegraphics[width=1in,height=1.25in, clip, keepaspectratio]{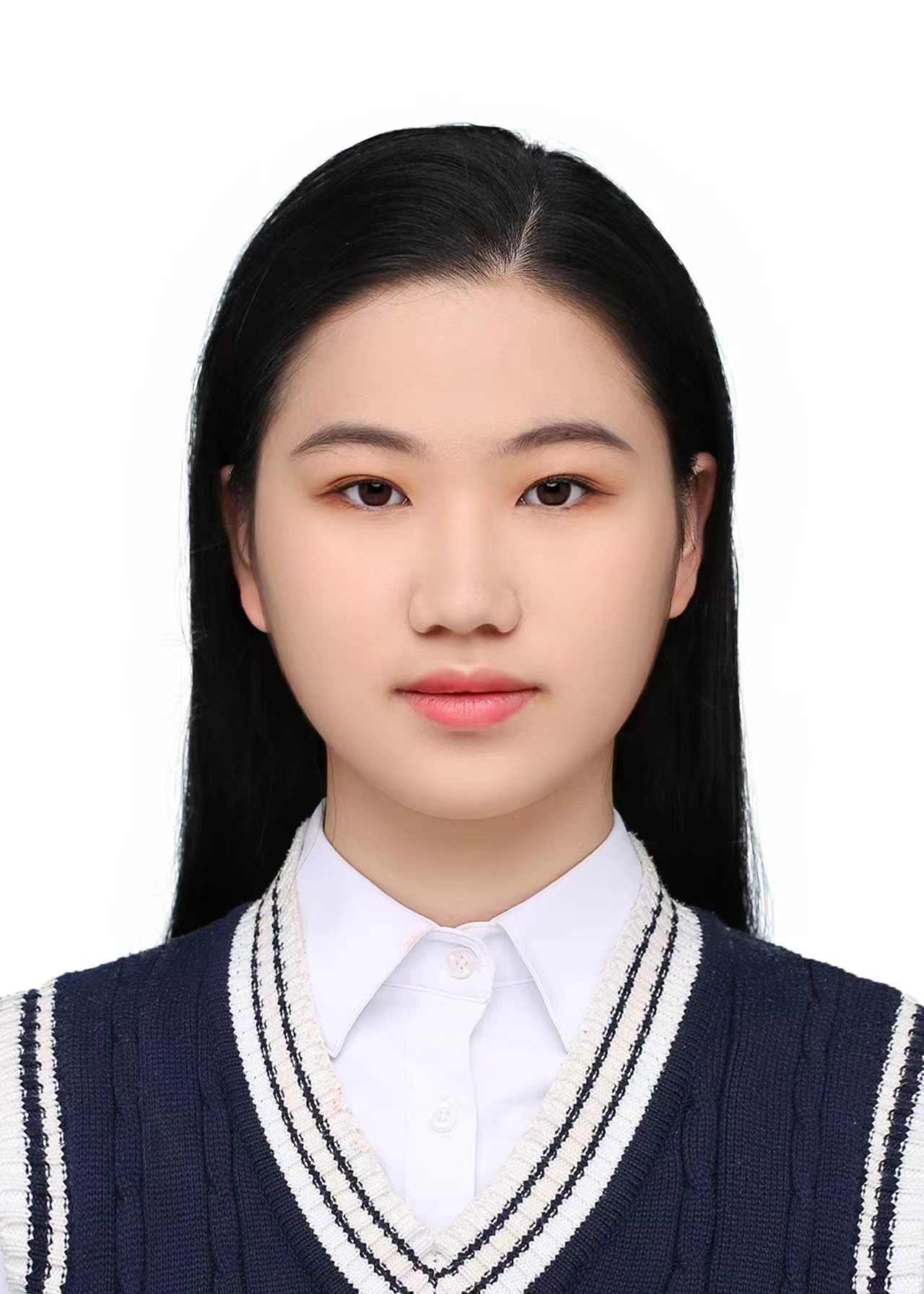}}]{Shuqi Shen} is an undergraduate student majoring in measurement and control technology and instrumentation at the University of Science and Technology Beijing, and will receive B.S. degree in 2025. She is currently a research assistant in Intelligent Transportation at the Hong Kong University of Science and Technology (Guangzhou), under the supervision of Prof. Meixin Zhu. Her main research focus is deep learning and reinforcement learning, involving multiple fields such as computer vision and autonomous driving.

\end{IEEEbiography}

\begin{IEEEbiography}[{\includegraphics[width=1in,height=1.25in, clip, keepaspectratio]{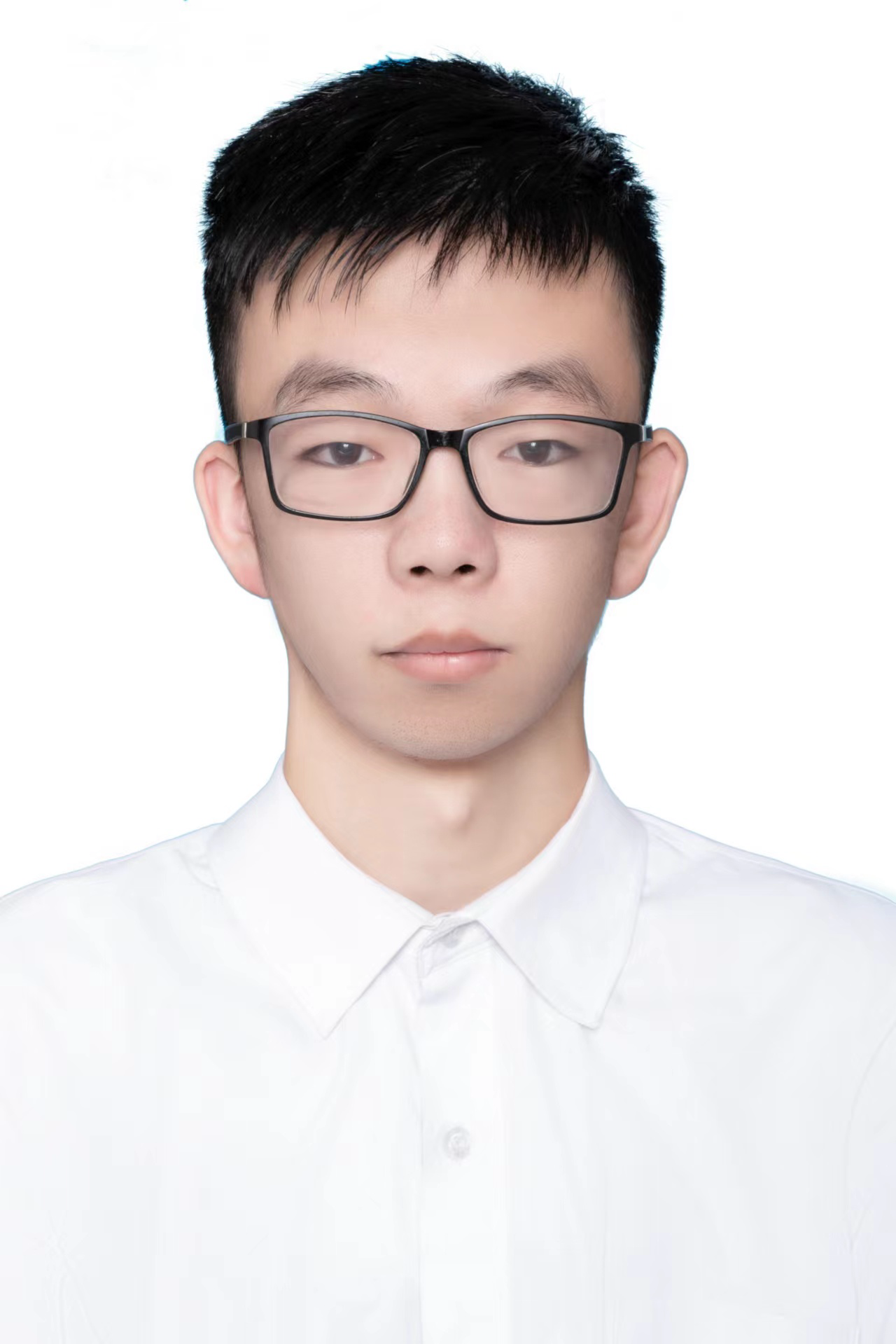}}]{Junjie Yang} is an undergraduate student majoring in measurement and control technology and instrumentation at the University of Science and Technology Beijing, and will receive B.S. degree in 2024. He is currently a research assistant and offer holder in Intelligent Transportation at the Hong Kong University of Science and Technology (Guangzhou), under the supervision of Prof. Meixin Zhu. His research interests include Autonomous Driving Decision Making and Planning, Pattern Recognition and Deep Learning, and Driving Behavior Modeling.
\end{IEEEbiography}

\end{document}